\documentclass[conference]{IEEEtran}%
\IEEEoverridecommandlockouts
\usepackage{cite}
\usepackage{footmisc}
\usepackage{amsmath,amssymb,amsfonts}
\usepackage{graphicx}
\usepackage{textcomp}
\usepackage{xcolor}
\usepackage{mathtools} %
\usepackage[textwidth=2cm]{todonotes} 
\usepackage[caption=false]{subfig} %
\usepackage[mathscr]{euscript} 
\usepackage{url} %
\usepackage{hyperref}
\usepackage{bm}
\usepackage{algpseudocode} 
\usepackage{algorithm}
\usepackage{xifthen}
\usepackage[export]{adjustbox}
\usepackage{textcomp}
\usepackage{array}   
\usepackage{svg}
\usepackage{enumitem}
\usepackage{xcolor}
\usepackage{booktabs}
\usepackage{enumitem}
\usepackage{todonotes}
\usepackage{xcolor}
\usepackage{tcolorbox}
\usepackage{subcaption}
\usepackage{caption}

\usepackage{wrapfig} 
\usepackage{colortbl}  
\newcommand{\mycomment}[1]{}
\NewDocumentCommand{\vect}{ O{} O{} m }{\mathbf{#3}\ifthenelse{\isempty{#1}}{}{^{(#1)}}\ifthenelse{\isempty{#2}}{}{_{#2}}}
\NewDocumentCommand{\mat}{ O{} O{} m }{\mathbf{#3}\ifthenelse{\isempty{#1}}{}{^{(#1)}}\ifthenelse{\isempty{#2}}{}{_{#2}}}
\NewDocumentCommand{\ten}{ O{} O{} m }{\pmb{\mathscr{#3}}\ifthenelse{\isempty{#1}}{}{^{(#1)}}\ifthenelse{\isempty{#2}}{}{_{#2}}}

\def\BibTeX{{\rm B\kern-.05em{\sc i\kern-.025em b}\kern-.08em
    T\kern-.1667em\lower.7ex\hbox{E}\kern-.125emX}}
\definecolor{mygreen}{rgb}{0,0.6,0}
\definecolor{mymauve}{rgb}{0.58,0,0.82}
\definecolor{mygray}{rgb}{0.5,0.5,0.5}
\definecolor{blue}{rgb}{0.0,0.0,1.0}
\definecolor{red}{rgb}{1.0,0.0,0.0}
\usepackage{listings}
\lstdefinelanguage{cypher2}{
    sensitive=true,
    morekeywords=[1]{MATCH, RETURN, WHERE, CONTAINS},
    morekeywords=[2]{PERSON, FRIEND,Document, Keyword, Affiliation, Country},
    morestring=[b]",
    morecomment=[l]{//},
    morecomment=[s]{/*}{*/},
    morecomment=[s]{--}{\ },
}
\lstdefinestyle{cypherstyle2}{
    language=cypher2,
    basicstyle=\footnotesize\ttfamily,
    keywordstyle=\color{blue}\bfseries, 
    keywordstyle=[2]\color{red}\bfseries,   
    commentstyle=\color{mygreen},
    stringstyle=\color{mymauve},
    numberstyle=\tiny\color{mygray},
    breaklines=true,
    showstringspaces=false,
    captionpos=b
}

\usepackage{listings}
\begin{document}
\title{Detecting Satellites in Radio-Frequency Data via Semi-Supervised Learning}

\author{\IEEEauthorblockN{
Cade W. Trotter\IEEEauthorrefmark{1},
Maksim E. Eren\IEEEauthorrefmark{2},
Justin C. Holmes\IEEEauthorrefmark{3},\\
J. Brent Parham\IEEEauthorrefmark{4},
David Ewing\IEEEauthorrefmark{4},
Boian S. Alexandrov\IEEEauthorrefmark{5},
and
Gian Luca Delzanno\IEEEauthorrefmark{3}
}
\IEEEauthorblockA{
\IEEEauthorrefmark{1}Modeling and Observations of Earth Systems, Los Alamos National Laboratory, Los Alamos, New Mexico, USA. \\
\IEEEauthorrefmark{2}Computational Intelligence \& Modeling, Los Alamos National Laboratory, Los Alamos, New Mexico, USA. \\
\IEEEauthorrefmark{3}Applied Mathematics and Plasma Physics, Los Alamos National Laboratory, Los Alamos, New Mexico, USA.\\
\IEEEauthorrefmark{4}MIT Lincoln Laboratory, Lexington, Massachusetts, USA.\\
\IEEEauthorrefmark{5}SLIC.AI, Santa Fe, New Mexico, USA.
}

\thanks{U.S. Government work not protected by U.S. copyright.

This research is based upon work supported in part by the Office of the Director of National Intelligence (ODNI), Intelligence Advanced Research Projects Activity (IARPA). The views and conclusions contained herein are those of the authors and should not be interpreted as necessarily representing the official policies, either expressed or implied, of ODNI, IARPA, or the U.S. Government. The U.S. Government is authorized to reproduce and distribute reprints for governmental purposes notwithstanding any copyright annotation therein.}
}

\maketitle

\begin{abstract}
Radio-frequency (RF) monitoring is crucial for maintaining space domain awareness, but frequently produces large, unlabeled, and variable datasets whose detections of satellites and space debris is often sparse. RF monitoring provides observations of satellites, space debris, and the background ionospheric environment, but can require specialized subject matter expertise to determine the nature of each observation. Existing deep learning techniques have shown success when applied to large labeled datasets, but they typically require many annotated examples and careful retraining to generalize across changing RF conditions. In contrast to the current supervised learning approaches, semi-supervised modalities can address data availability and transferability issues in limited-data RF settings because it can use unlabeled observations to detect latent patterns interpretable by subject matter experts. In this paper, we apply a semi-supervised RF detection and classification workflow for satellite detection, that combines Non-negative Matrix Factorization with automatic model determination (NMFk), subject-matter expert supervised interpretation, and a classifier based prediction from our semi-supervised decision making. We first represent RF observations as a non-negative feature matrix and use NMFk to estimate the ideal number of clusters to represent the patterns in our unlabeled RF data. Subject-matter expertise is then applied to the clusters to assign physical meaning, such as satellite detections, basic ionospheric environmental conditions, or other RF event categories. We then train an XGBoost classifier on the interpretable clustering to test on a reserved test-set and categorize future data. This pipeline reduces dependence on large pre-labeled datasets by using unsupervised factorization and subject matter expertise to establish an interpretable and applicable methodology for detecting, observing, and classifying behavior in RF data.

\end{abstract}


\section{Introduction}
\label{sec:introduction}
The increased commercialization of space over the past decade has intensified the potential risks of orbiting debris prompting the need for greater space domain awareness ~\cite{bennett2025orbital}. Radio-frequency (RF) sensing provides an important source of situational awareness for monitoring satellites, debris, and anomalous emissions. However, RF observational datasets are often dynamic, and can lack crucial labeling of the physical phenomena contained in them. The need for physically interpretable RF data in event classification presupposes datasets with known structure such as recurring satellite passes, and background ionospheric environmental conditions. This makes RF event classification difficult in practice: a model must be supplied with observations and activity representative of the phenomena they wish to classify and separate from background behavior while remaining robust to new patterns that may arise in future data.

Fully supervised learning can be applied to RF sensing where large, representative labeled datasets are available, but the required data is often unavailable for RF monitoring applications. You only look once (YOLO) style \cite{Redmon2015YouOL} deep learning approaches have found success in the characterization and detection of events, but require careful pre-labeling and even retraining in the case where environmental or observational conditions change \cite{zhu2025cross}. In the context of RF monitoring, predefined labeling of spectral data would require tremendous subject-matter expert review and datasets may not have enough space object detections to support the pattern recognition of important phenomena. Additionally, supervised methods may fail to generalize to changing background RF conditions. 

Semi-supervised learning offers a practical solution for RF monitoring datasets that lack labeling. Rather than requiring subject matter expert (SME) intensive labels, semi-supervised workflows can use unlabeled data to characterize structure and depict latent patterns that can be ascribed meaning by SMEs with reduced effort. This is especially useful for RF event analysis because large volumes of unlabeled observations are often available, while expert labels are comparatively scarce. In this paper, we propose a semi-supervised approach of clustering representative latent patterns in the RF data, applying subject matter expertise (SME) to decipher the physical meaning of each cluster, then train a supervised model on to predict/assign RF samples into the SME tagged clusters. 

Specifically, our approach combines Non-negative Matrix Factorization with automatic model determination (NMFk\footnote{NMFk is in a publicly available GitHub repository named T-ELF\cite{TELF} we accessed at: \url{https://github.com/lanl/T-ELF}}) \cite{eren2023semi, alexandrov2020patent}, SME-based cluster interpretation, and XGBoost \cite{10.1145/2939672.2939785} classification. We first represent RF observations as a non-negative feature matrix and use NMFk to estimate the ideal number of clusters to represent the patterns in our unlabeled RF data. Subject-matter expertize is then applied to the clusters to assign physical meaning, such as satellite detections, basic ionospheric environmental conditions, or other RF event categories. We then train a XGBoost classifier on the interpretable clustering to test on a reserved test-set and categorize future data. In summary, our contributions include:
\begin{enumerate}
    \item Applying a semi-supervised NMFk-based classification framework to RF data for space domain situational awareness.
    \item Demonstrating how NMFk can be used on unlabeled RF observations to estimate latent and recurring physical patterns in RF data.
    \item Training an XGBoost classifier on SME-interpreted NMFk cluster labels to assign new RF samples to the discovered event categories (i.e., clusters).
    \item Validating the NMFk-based detections and classifier predictions against TLE-derived ground truth to assess how well the discovered RF event categories correspond to known tracked space objects.
\end{enumerate}

\section{Related Works}
\label{sec:related_works}
Space domain awareness is an expanding domain with a well established body of work in both radar and optical detection modalities \cite{national2011limiting}. Extensive work has been done in optical detection of space objects \cite{schildknecht2007optical, hickson2018fast}. While highly effective for many use cases, optical detection modalities suffer from dependence on solar illumination both from a time of day and signal perspective as well as the potential for weather obfuscation of object detection. Radar based systems address many of these challenges and have found success in the detection of debris and satellites, but are costly to operate \cite{foster2005detection, mehrholz2002detecting}. Applications of traditional signal to noise analysis and thresholding are well-established and have yielded good results \cite{prabu2020development}, but can miss signatures and patterns identified in more complex techniques. Improvements in artificial intelligence (AI) and machine learning (ML) techniques have bolstered the use of deep learning approaches to object detection through YOLO based object detection schemes \cite{massimi2024deep, zhang2025semi}. These approaches have demonstrated great potential for object detection, but rely on large labeled datasets which are currently scarce. Use of synthetic radar data \cite{massimi2024deep} has been successful in the YOLO framework, but may face challenges in data availability for large labeled observational datasets. NMFk with automatic model determination has been shown to be a reliable dimensionality reduction method for latent feature extraction and clustering in unlabeled data \cite{eren2023semi,alexandrov2020patent}, making it well suited for clustering of unlabeled RF samples.

 


 \section{Methods}
\label{sec:methods}
\begin{table*}[t]
\centering
\small
\caption{Notation conventions used throughout this paper.}
\label{table:notations}
\begin{tabular}{c|l||c|l}
\hline
\textbf{Notation}  & \textbf{Description}  & \textbf{Notation}  & \textbf{Description}  \\ \hline
$x$                & Scalar                 & $\vect{x}_i$       & $i$th entry of vector $\vect{x}$\\
$\vect{x}$         & Vector                 & $\mat{X}_{ij}$     & Entry in row $i$ and column $j$\\
$\mat{X}$          & Matrix                 & $\mat{X}_{i:}$     & $i$th row of $\mat{X}$\\
$\ten{X}$          & Tensor                 & $\mat{X}_{:j}$     & $j$th column of $\mat{X}$\\
$\ten{X}_{::i}$    & $i$th slice along the third mode  & $\ten{X}_{:::i}$   & $i$th slice along the fourth mode\\
$\ten{X}^{\textit{name}}$  & Superscript identifier & $*$ & Matrix product\\
$\odot$           & Element-wise product    & $\otimes_B$ & Boolean matrix multiplication\\
\hline
\end{tabular}
\end{table*}

This section first describes the RF dataset used in our analysis. In this paper, we apply NMFk  \cite{alexandrov2020patent, eren2023semi} for space domain awareness; therefore, for completeness we include brief description of the method. Then, this section describes the SME cluster interpretation, and downstream XGBoost classification workflow used in this study. For convenience, the notation used throughout this section is summarized in Table~\ref{table:notations}.

\subsection{Data \& Preparation}
\label{sec:dataprep}

\begin{table}[!t]
\centering
\small
\caption{Data description of the data used in this paper. The data consists of 3hrs of radar spectra with sparse detections of satellites. Throughout the paper the data is referenced in various ways: train, test, and total count of these values. This table serves as a reference for understanding what split of the data is being used at any point in our analysis.}
\label{nmfk_data_table}
\begin{tabular}{c|l|l|l}
\hline
\textbf{Data Info.} & \textbf{Train} & \textbf{Test} & \textbf{Total}  \\ \hline
Percentage (\%)        & 80 & 20 & 100 \\
Spectra Samples        & 2181 & 546 & 2727 \\
Time (hrs)             & $\approx 2.4$ & $\approx 0.6$ & 3 \\
Known Starlink Satellites             & 34 & 12 & 46 \\
Integration Time (s) / sample     & 4 & 4  & 4 \\

\hline

\end{tabular}
\end{table}

The data used in this analysis are in-phase and quadrature (I/Q) measurements from the ARPA Long-Range Tracking and Instrumentation Radar (ALTAIR) \cite{Stone2000} on Kwajalein Atoll. The data were collected on November 15, 2024, for the Space Debris Identification and Tracking (SINTRA) program \cite{sintra}. This dataset is intended to support the identification of spacecraft and orbital debris with very small radar cross sections. The selected data were collected at a radar operating frequency of 158 MHz with a full receiving bandwidth of 156 kHz.

For each 0.05-second radar pulse interval, we processed the I/Q data by convolving the received signal with the transmitted uncoded pulse for each range gate and then applying a Fourier transform to convert the data to the frequency domain. We summed the resulting spectra over 80 pulses, corresponding to 4 seconds of integration, to resolve the so-called ``ion line,'' a feature generated by incoherent Thomson scattering of the radar pulse from the ionosphere \cite{Dougherty1961,Hysell2020}. A useful review of incoherent scatter radar features is provided by \cite{Akbari2017}. The final output is a time series of range--Doppler power spectra, with examples shown in Figure~\ref{cluster_spectra}. Objects passing through the radar beam appear as bright features at ranges and Doppler shifts corresponding to their distance and radial velocity relative to the radar dish. Identifying these features and associating them with different classes of objects is the focus of this study.

Ground-truth metadata were constructed by propagating two-line elements (TLEs), which define the satellite orbits of known objects, through the radar beam. The TLE data were retrieved from \url{www.space-track.org}, a service provided by USSPACECOM \cite{spacetrack}. TLEs were propagated using the \texttt{sgp4} Python library and converted to azimuth, elevation, and range relative to the radar beam using the \texttt{astropy} Python library. Although the radar beam width is 2.9 degrees at 158 MHz, large objects passing through the side lobes of the radar beam may also be detected. Therefore, for each object passing within 5.31 degrees of the beam center, we recorded the timestamp, NORAD ID, range, and radial velocity.

One caveat is that the TLE-derived metadata are extensive but not complete. The data used here are General Perturbations positional data from the space catalog for 2023 and 2024. To estimate catalog completeness, we compared these data with the \url{celestrak.org} SATCAT catalog of objects, retrieved on April 13, 2026 \cite{celestrak}. This catalog contains 7,272 Starlink satellites launched before November 15, 2024, whereas the historical TLE data used in this study contain 7,014 of these objects. All missing Starlink satellites were launched before 2023, so some fraction may have deorbited. In addition, TLEs become less accurate when propagated over long time periods, so older TLE entries may have incorrect positions. Based on this comparison, we estimate the dataset to be more than 95\% complete.

Using the TLE propagation metadata, we predict 72 total Starlink satellite detections by ALTAIR within the dataset time range. The full 3-hour, preprocessed, NMFk-ready ALTAIR RF dataset is summarized in Table~\ref{nmfk_data_table}. 

\subsection{Non-negative Matrix Factorization}
\label{subsection:nmf}

NMFk is a Non-negative Matrix Factorization (NMF) method, with additional capability to estimate the number of clusters $k$. NMF is a low-rank representation method for non-negative data. Given an observation matrix
$\mat{X}\in \mathbb{R}_{+}^{n \times m}$, NMF seeks two non-negative factor matrices whose product approximates the original data,
\begin{equation}
    \mat{X} \approx \hat{\mat{X}} = \mat{W}\mat{H},
\end{equation}
where $\mat{W}\in \mathbb{R}_{+}^{n\times k}$, $\mat{H}\in \mathbb{R}_{+}^{k\times m}$, and $k \ll \min(n,m)$. In this formulation, $n$ is the number of measured features, $m$ is the number of RF samples, and $k$ is the latent dimension of the approximation. The columns of $\mat{W}$ represent non-negative basis patterns, while the columns of $\mat{H}$ describe how strongly each basis pattern contributes to each sample.

The factorization is obtained by minimizing the Frobenius-norm reconstruction loss subject to non-negativity constraints:
\begin{align}
\label{eqn:nmf_min}
    \underset{\mat{W} \in \mathbb{R}_{+}^{n\times k}, \, \mat{H} \in \mathbb{R}_{+}^{k \times m}}{\operatorname{min}}
    \; \|\mat{X} - \mat{W}\mat{H}\|_{F}^{2}.
\end{align}
Although this objective is non-convex in $\mat{W}$ and $\mat{H}$ jointly, it can be optimized through alternating updates. We use multiplicative updates, which maintain non-negativity throughout the optimization \cite{lee1999learning}. Under the Frobenius objective, the update rules are
\begin{align}
\label{eqn:NMF_updates_W}
\mat{W} &\leftarrow \mat{W}\odot
\frac{\mat{X}\mat{H}^T}{\mat{W}\mat{H}\mat{H}^T},
\\
\label{eqn:NMF_updates_H}
\mat{H} &\leftarrow \mat{H}\odot
\frac{\mat{W}^T\mat{X}}{\mat{W}^T\mat{W}\mat{H}}.
\end{align}
The quality of the approximation is measured using the relative reconstruction error,
\begin{align}
\label{eqn:rel_error}
\text{Relative Error}
= \frac{\|\mat{X}-\mat{W}\mat{H}\|_F^2}{\|\mat{X}\|_F^2}.
\end{align}
With the Frobenius loss, NMF also admits a probabilistic interpretation related to Gaussian noise assumptions \cite{fevotte2009nonnegative}.

\subsection{Non-negative Matrix Factorization with Automatic Model Determination}
\label{subsection:nmfk}

A key limitation of standard NMF is that the rank $k$, or the number of clusters, must be selected before factorization. If $k$ is too small, the model cannot capture the dominant structure in $\mat{X}$ and underfits the observations. If $k$ is too large, the learned factors can become fragmented, unstable, or dominated by noise, which reduces interpretability. Thus, selecting $k$ is a model-determination problem: the objective is to find a latent dimension that reconstructs meaningful structure while remaining stable across perturbations of the data.

Several approaches have been proposed for estimating latent dimensionality. Automatic Relevance Determination (ARD) was introduced in Bayesian modeling \cite{mackay1994bayesian}, adapted to Principal Component Analysis \cite{bishop1999bayesian}, and later extended to NMF \cite{morup2009tuning,tan2012automatic}. Another class of approaches estimates $k$ by examining the stability of repeated NMF solutions, based on the premise that meaningful latent factors should persist across related factorizations of perturbed data \cite{brunet2004metagenes}.

In this work, we use NMFk, an NMF-based model-determination procedure that combines reconstruction accuracy, solution stability, and clustering statistics to estimate the number of latent factors \cite{alexandrov2020patent}. NMFk has been used in large-scale genomic studies \cite{alexandrov2013signatures,alexandrov2020repertoire,islam2022uncovering}, evaluated on synthetic datasets with known latent structure \cite{nebgen2021neural}, applied to topic extraction problems where it compared favorably with alternatives such as spherical $k$-means \cite{vangara2020semantic}, and used for large scale malware family classification \cite{eren2023semi}. The method repeatedly perturbs the input matrix, performs NMF over a range of candidate ranks, clusters the resulting factors, and selects a rank that balances reconstruction accuracy with cluster stability as measured by Silhouette statistics \cite{ROUSSEEUW198753}.

Algorithm~\ref{alg:nmfk} summarizes the NMFk procedure.

\begin{algorithm}[htb]
    \caption{NMFk($\mat{X}$, $k^{min}$, $k^{max}$, $M$, $Sill_{thr}$)}
    \label{alg:nmfk}
    \begin{algorithmic}[1]
        \Require Non-negative matrix $\mat{X} \in \mathbb{R}_{+}^{n \times m}$, candidate ranks $k^{min}$ to $k^{max}$, number of perturbations $M$, stability threshold $Sill_{thr}$
        \For{$k = k^{min}$ to $k^{max}$}
            \For{$q = 1$ to $M$}
                \State $\ten{X}_{::q} \gets \textsc{Perturb}(\mat{X})$
                \State $(\ten{W}_{::kq}, \ten{H}_{::kq}) \gets \textsc{NMF}(\ten{X}_{::q}, k)$
            \EndFor
            \State $\ten{W}^{all} \gets [\ten{W}_{::k1}, \ldots, \ten{W}_{::kM}]$
            \State $\ten{H}^{all} \gets [\ten{H}_{::k1}, \ldots, \ten{H}_{::kM}]$
            \State $(\ten{\hat{W}}, \ten{\hat{H}}) \gets \textsc{CustomCluster}(\ten{W}^{all}, \ten{H}^{all})$
            \State $\ten{\widetilde{W}}_{::k} \gets \textsc{ClusterMedians}(\ten{\hat{W}})$
            \State $\ten{H}^{reg}_{::k} \gets \textsc{NNLS}(\mat{X}, \ten{\widetilde{W}}_{::k})$
            \State $\vect{s}_{k} \gets \textsc{SilhouetteScores}(\ten{\hat{W}})$
            \State $\vect{err}_{k} \gets \textsc{ColumnErrors}(\mat{X}, \ten{\widetilde{W}}_{::k}, \ten{H}^{reg}_{::k})$
        \EndFor
        \State $\vect{err}^{all} \gets [\vect{err}_{k^{min}}, \ldots, \vect{err}_{k^{max}}]$
        \State $k^{opt} \gets \textsc{LStatistics}(\vect{err}^{all}, \vect{s}_{k}, Sill_{thr})$
        \State \Return $\ten{\widetilde{W}}_{::k^{opt}}$, $\ten{H}^{reg}_{::k^{opt}}$, $k^{opt}$
    \end{algorithmic}
\end{algorithm}

The main steps are as follows.

\begin{enumerate}
    \item \textit{Perturbation ensemble.}
    NMFk constructs an ensemble of $M$ perturbed versions of the original matrix. Each perturbed matrix has the same expected value as $\mat{X}$, but its entries are modified by small random noise:
    \begin{equation}
        \ten{X}_{ijq} = \mat{X}_{ij} + \delta_{ijq},
    \end{equation}
    where $\delta_{ijq}$ is a small perturbation sampled independently for entry $(i,j)$ and replicate $q$.

    \item \textit{Repeated NMF over candidate ranks.}
    For each candidate rank $k \in [k^{min}, k^{max}]$, NMF is applied to every perturbed matrix. This produces an ensemble of $M$ factor pairs,
    \begin{equation}
        [\ten{W}_{::kq}, \ten{H}_{::kq}]_{q=1}^{M}.
    \end{equation}
    In this study, the underlying NMF solver uses Frobenius-norm multiplicative updates \cite{lee1999learning}.

    \item \textit{Factor clustering.}
    For a fixed $k$, the columns of the $M$ estimated $\mat{W}$ matrices are pooled and clustered. The NMFk clustering procedure is constrained so that each cluster contains one latent factor from each of the $M$ NMF runs. This constraint reflects the fact that each replicate produces exactly $k$ factors. Cosine similarity is used to compare columns during clustering.

    \item \textit{Robust factor construction.}
    After clustering, each cluster is summarized by its median factor, yielding a robust basis matrix $\ten{\widetilde{W}}_{::k}$. The corresponding coefficient matrix $\ten{H}^{reg}_{::k}$ is then recomputed by non-negative least squares regression of $\mat{X}$ onto $\ten{\widetilde{W}}_{::k}$.

    \item \textit{Stability assessment.}
    Cluster stability is evaluated using Silhouette statistics \cite{ROUSSEEUW198753}. Silhouette scores quantify how well each factor belongs to its assigned cluster relative to neighboring clusters. Scores range from $-1$ to $1$, with larger values indicating more coherent and better separated clusters.

    \item \textit{Reconstruction assessment.}
    For each candidate rank, NMFk measures how accurately the robust factors reconstruct the input data:
    \begin{equation}
        R_k =
        \frac{\|\mat{X} - \ten{\widetilde{W}}_{::k}\ten{H}^{reg}_{::k}\|_F}
        {\|\mat{X}\|_F}.
    \end{equation}
    Column-wise reconstruction errors are also retained:
    \begin{equation}
        e_{jk} =
        \frac{\|\mat{X}_{:j} - \ten{X}^{rec}_{:jk}\|}
        {\|\mat{X}_{:j}\|},
        \quad j = 1,\ldots,m.
    \end{equation}

    \item \textit{Rank selection with L-statistics.}
    NMFk compares the distributions of column-wise reconstruction errors across candidate ranks using L-statistics \cite{vangara2021finding}. A two-sided Wilcoxon rank-sum test is used to evaluate whether the reduction in reconstruction error remains statistically meaningful as $k$ increases \cite{Haynes2013}. The selected rank $k^{opt}$ is the largest rank that maintains stable clusters, according to the minimum Silhouette threshold, while avoiding additional factors that primarily fit noise. In our experiments, we use a minimum Silhouette threshold of $0.80$ to favor stable and interpretable RF clusters.
\end{enumerate}

\subsection{Cluster Interpretation and Semi-supervised RF Event Classification}
\label{subsection:cluster_use}

After NMFk heuristically estimates the optimal rank $k^{opt}$ and returns the robust factors
$\ten{\widetilde{W}}_{::k^{opt}}$ and $\ten{H}^{reg}_{::k^{opt}}$, we use the resulting latent representation to construct event labels and train a classifier for future RF samples. The columns of $\ten{\widetilde{W}}_{::k^{opt}}$ represent recurring RF patterns discovered from the data, while each column of $\ten{H}^{reg}_{::k^{opt}}$ gives the activation strength of those patterns for a particular sample.

We assign each training sample to its dominant NMFk component using the largest activation in its coefficient vector. For sample $j$, the cluster assignment is
\begin{equation}
\label{eqn:nmfk_cluster_assignment}
    c_j = \operatorname*{arg\,max}_{\ell \in \{1,\ldots,k^{opt}\}} \ten{H}^{reg}_{\ell j}.
\end{equation}
This produces a discrete cluster assignment while retaining the continuous activation profile that describes how strongly each latent RF pattern is expressed.

The NMFk clusters are then reviewed by subject-matter experts. This review step is necessary because matrix factorization discovers recurring structure, but it does not by itself assign operational meaning to each component. SMEs inspect representative samples from each cluster, including high-activation examples, temporal context, RF signatures, and any available metadata. Based on this review, each cluster is assigned a semantic event label, such as satellite pass, no-event/background activity, or another RF event category. In this way, the NMFk output provides data-driven candidate classes, and SME interpretation converts those classes into labels that are meaningful for RF analysis.

This workflow is semi-supervised rather than purely supervised. The initial cluster structure is learned from unlabeled RF samples through NMFk, so the method does not require event labels before discovery. Human supervision is introduced after clustering, when SMEs attach semantic labels to the discovered groups. These SME-interpreted cluster labels are then used as training targets for a supervised classifier. This approach follows the general semi-supervised learning setting, where unlabeled data are used to exploit the structure of the input distribution and a smaller amount of labeled information is used to guide classification \cite{chapelle2006semi}.

Given the SME-interpreted label $y_j$ for each sample $j$, we train an XGBoost classifier \cite{10.1145/2939672.2939785} to map RF feature vectors to cluster-derived event labels. Let $\vect{z}_j$ denote the feature vector for sample $j$. The classifier learns a function
\begin{equation}
\label{eqn:xgb_cluster_classifier}
    f_{\theta}: \vect{z}_j \mapsto y_j,
\end{equation}
where $f_{\theta}$ is the trained XGBoost model. At inference time, a new RF sample is transformed into the same feature representation and passed to $f_{\theta}$ to predict its event category.

The resulting pipeline separates discovery, interpretation, and deployment. NMFk discovers stable RF patterns without requiring labels. SME review assigns meaning to the discovered clusters. XGBoost then provides a supervised model for assigning new samples to the SME-interpreted categories. This enables automated classification of future RF observations into satellite passes, background/no-event periods, or other events defined through expert review.

 \section{Results}
\label{sec:results}
\subsection{NMFk Results: Unsupervised Clusters and SME Decision Making}
\label{subsection: nmfk results}
\begin{figure*}[t]
  \centering
  \includegraphics[
    width=\textwidth,
    height=0.4\textheight,
    keepaspectratio
  ]{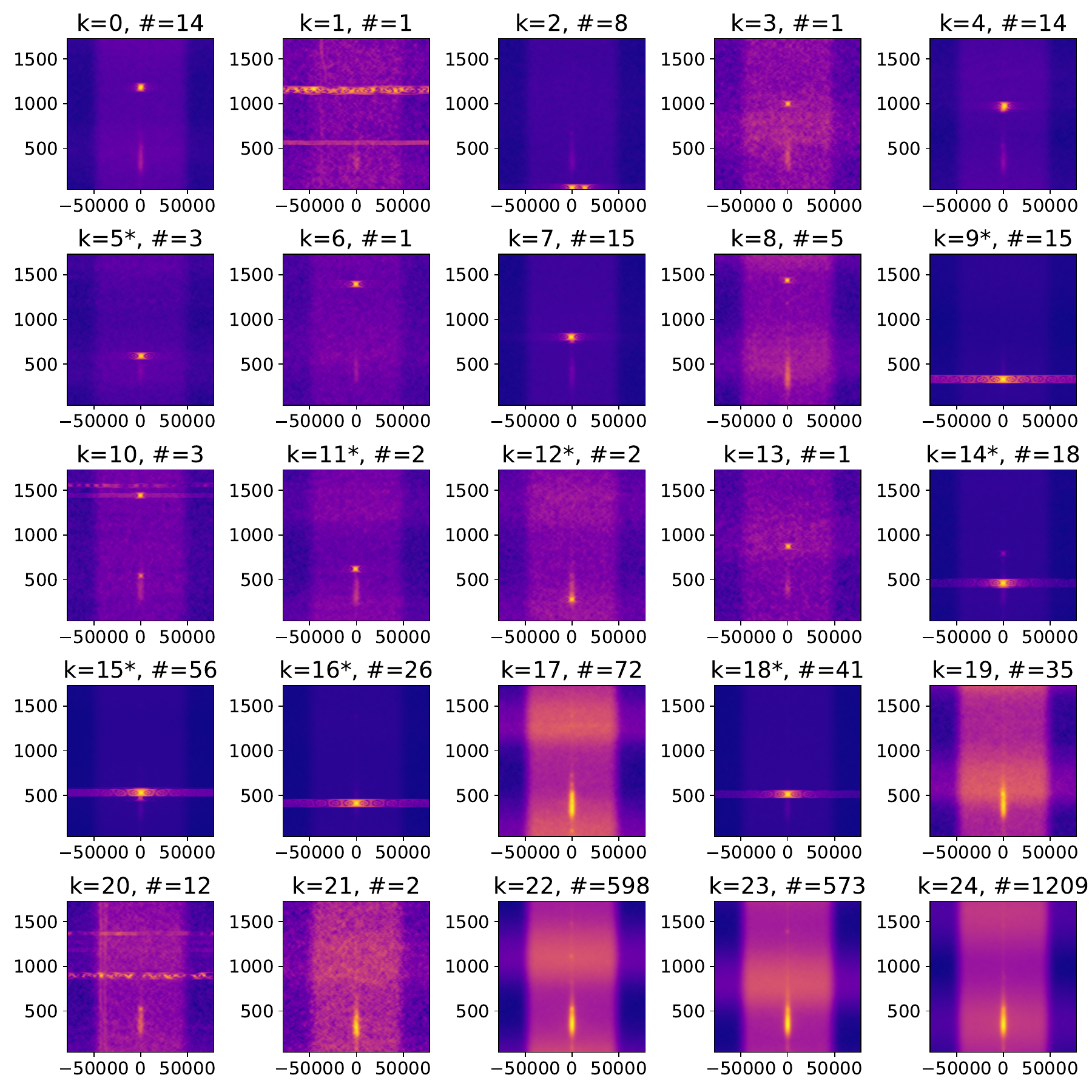}
  \caption{Spectral average for each cluster (k) in our NMFk analysis. The y axes of each subplot is range expressed in km, and the x axes are Doppler-frequency in Hz. Each cluster 0-24 is labeled and includes the number (\#) of samples within the cluster used to create that spectral average. The $*$ denotes that a cluster has been determined to be a ``Starlink'' cluster that we use for analysis in our results section. The \# denotes total number in each cluster for all samples (train and test) see Table \ref{nmfk_data_table}.}
  \label{cluster_spectra}
\end{figure*}

Running NMFk on the ALTAIR data set determined that the optimal number of representative clusters is 25. Each cluster 0-24 represents signatures in the spectra that can characterize a subset of the ALTAIR data. Our team then determined the nature of each cluster by analyzing aspects of each spectra such as range, Doppler frequency, minimum, maximum, and average values. Pixel-wise averages across all spectra in each cluster can be observed in Figure~\ref{cluster_spectra} where the cluster name, number of samples, and distinction of containing Starlink satellites ($*$ in title) can be observed. For the rest of the text, the word ``Starlink'' in quotations will refer to our clusters in which we have determined the features to be consistent with known Starlink satellites (but aren't necessarily actual Starlink satellites). Use of the word Starlink without quotations will be in reference to known Starlink satellites. The Doppler-range plots shown in Figure ~\ref{cluster_spectra} represent signatures of each cluster created by NMFk. We normalized the color scale in each plot to show the respective patterns for each average spectra. Examples like cluster 22 show what is essentially a background ionospheric environment with no clear bright returns from satellites. Clusters with the $*$ were inspected by our SMEs and determined to be at an altitude consistent with the expected orbit of Starlink satellites (to be used in validation analysis further in our results section). Other bright detections of satellites at higher/lower altitude (such as cluster 0, 4, 7, etc.) were interpreted by our SMEs as containing non-Starlink satellites. In our XGBoost analysis, some clusters were so sparse that they did not have enough samples in both the test and train split that they were not considered in the analysis. Analysis plots and tables will appear to have missing clusters for this reason (further discussion in IV.B). All figures and tables with reference to cluster numbers are consistent and Figure~\ref{cluster_spectra} can serve as a good reference for the behavior of the samples in that cluster. 


\subsection{XGBoost Classifier: Train/Test Split and Assessment}
\label{subsection:altair data}
Due to the radar side lobes, wide beam, and 4 second integration time of the radar returns, there were several instances of individual objects appearing in back to back integrated spectra. We did not use several of the clusters in training and testing due to the small number of RF samples that were assigned to these clusters ($k=1,3,4,5,6,9,11,12,13$ were not used). To train an XGBoost classifier we applied the standard 80/20 test/train split ratio, but we chose to split chronologically instead of pseudo-randomly, ensuring that the nearly identical samples (from the side lobes, wide beam, and 4 sec. integration time) would not be unfairly represented in both the test and train data. Because of this test/train split, some clusters with enough support for a pseudo-random split do not have adequate representation in the final 20\% of the data are dropped as such ($k=4,5,6,9$ are dropped for this reason). A classification report of our classifier (shown in Table ~\ref{classification_report_table} ) demonstrated micro-averages of: 0.875 for precision, 0.870 for recall, and 0.872 for f1. 

\begin{table}[t!]
\centering
\caption{Classification report of the XGBoost classifier on the test split of the data (see Table \ref{nmfk_data_table} and Section \ref{subsection:altair data} for sample information).}
\label{classification_report_table}
\begin{tabular}{lcccc}
\toprule
k & precision & recall & f1-score & support \\
\midrule
0 & 1.000 & 0.667 & 0.800 & 3 \\
2 & 0.000 & 0.000 & 0.000 & 1 \\
7 & 0.000 & 0.000 & 0.000 & 1 \\
8 & 0.000 & 0.000 & 0.000 & 2 \\
10 & 0.000 & 0.000 & 0.000 & 1 \\
14 & 1.000 & 1.000 & 1.000 & 1 \\
15 & 1.000 & 0.750 & 0.857 & 12 \\
16 & 1.000 & 1.000 & 1.000 & 14 \\
17 & 0.400 & 0.111 & 0.174 & 18 \\
18 & 0.889 & 1.000 & 0.941 & 8 \\
19 & 0.000 & 0.000 & 0.000 & 8 \\
20 & 1.000 & 1.000 & 1.000 & 1 \\
22 & 0.870 & 0.898 & 0.884 & 127 \\
23 & 0.831 & 0.934 & 0.879 & 121\\
24 & 0.902 & 0.925 & 0.913 & 228 \\
micro avg & 0.875 & 0.870 & 0.872 & 546 \\
macro avg & 0.593 & 0.552 & 0.563 & 546 \\
weighted avg & 0.846 & 0.870 & 0.854 & 546+ \\
\bottomrule
\end{tabular}
\end{table}

\begin{figure}[htb]
    \centering

    \begin{minipage}{0.49\linewidth}
        \centering
        \includegraphics[width=\linewidth]{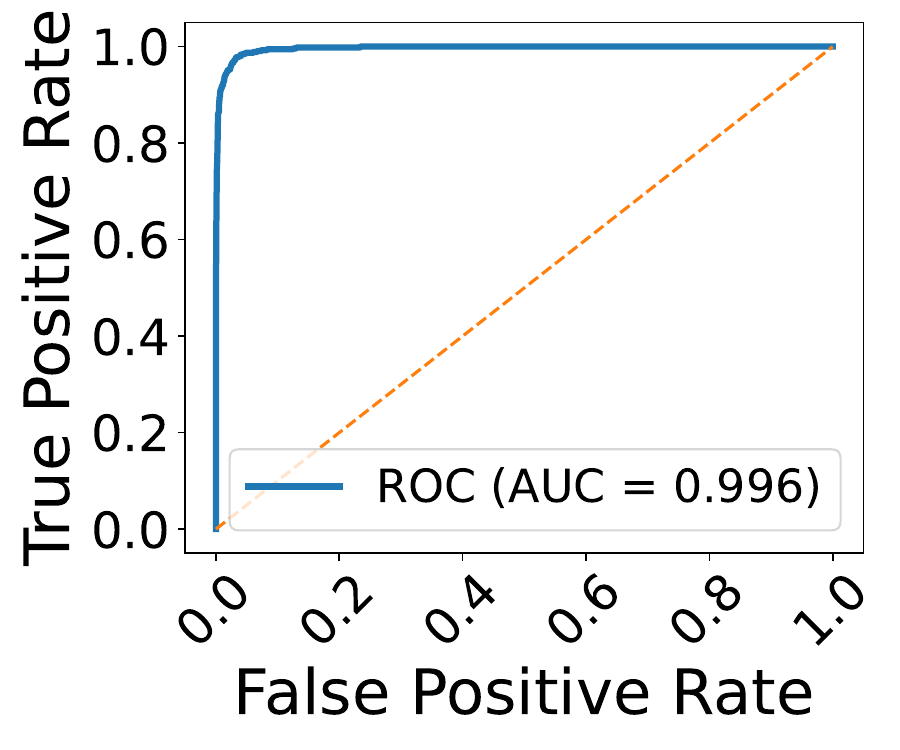}
        
        \small
        (a) ROC-AUC
        \label{fig:roc}
    \end{minipage}
    \hfill
    \begin{minipage}{0.49\linewidth}
        \centering
        \includegraphics[width=\linewidth]{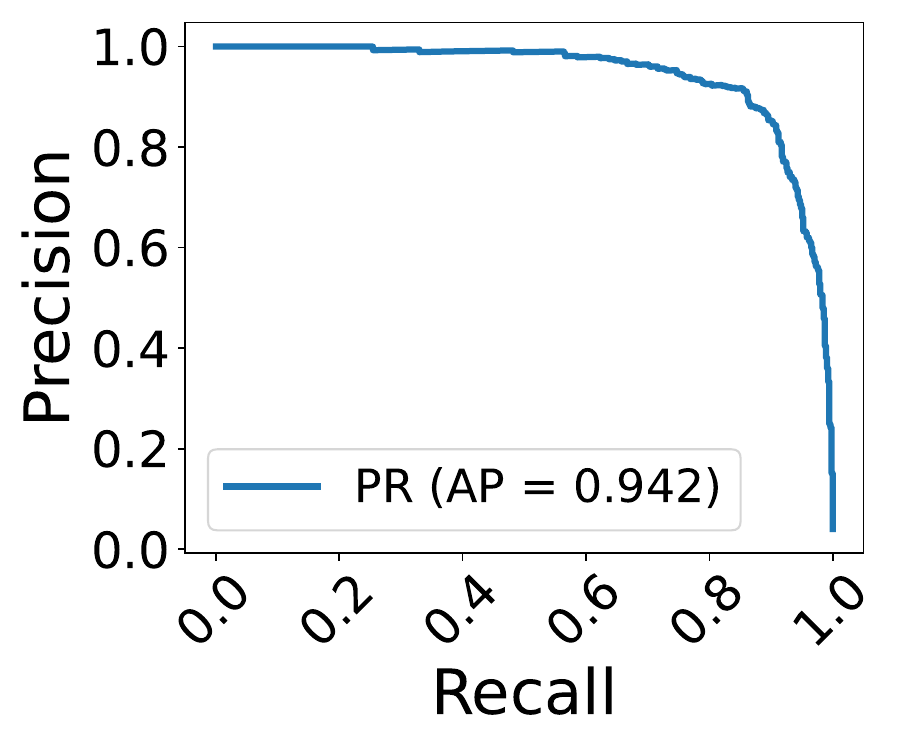}
        
        \small
        (b) PR-AUC
        \label{fig:pr}
    \end{minipage}

    \caption{Performance evaluation metrics for the XGBoost classifier. ROC (on the left) demonstrates our classifier's ability to distinguish between classes (clusters). For a purely random guess the ROC-AUC would be 0.5, where 1.0 would be a perfect guess. PR (on the right) is an overall assessment of our classifier. A mark of success is a retention of precision as recall increases (the curve being oriented to the top right).}
    \label{fig:xgb_metrics}
\end{figure}

Micro-average receiver operating characteristic area under the curve (ROC-AUC) and precision recall area under the curve (PR-AUC) plots (depicted in Figure~\ref{fig:xgb_metrics}) show micro-averages of 0.996 and 0.942 for ROC and PR respectively.

In contrast to the sample weighted micro-average metrics, the macro-average F1 score of 0.593 serves as a broader representation of our classifier's performance by equally weighting each cluster. Due to the nature of our chronological test/train split, several of our minority clusters had fewer than 10 samples resulting in patterns that could not reliably be learned by the classifier. We treat this as a known limitation of the test/train split rather than a failure of the underlying clustering. To address this in future work we will sample larger datasets that will include enough support for each minority cluster to meaningfully evaluated.

\subsection{Predictive Success: Satellite Detection and Classification}
\label{subsection:satellite detection}

\begin{figure}[htb]
    \centering
    \includegraphics[width=\columnwidth,keepaspectratio]{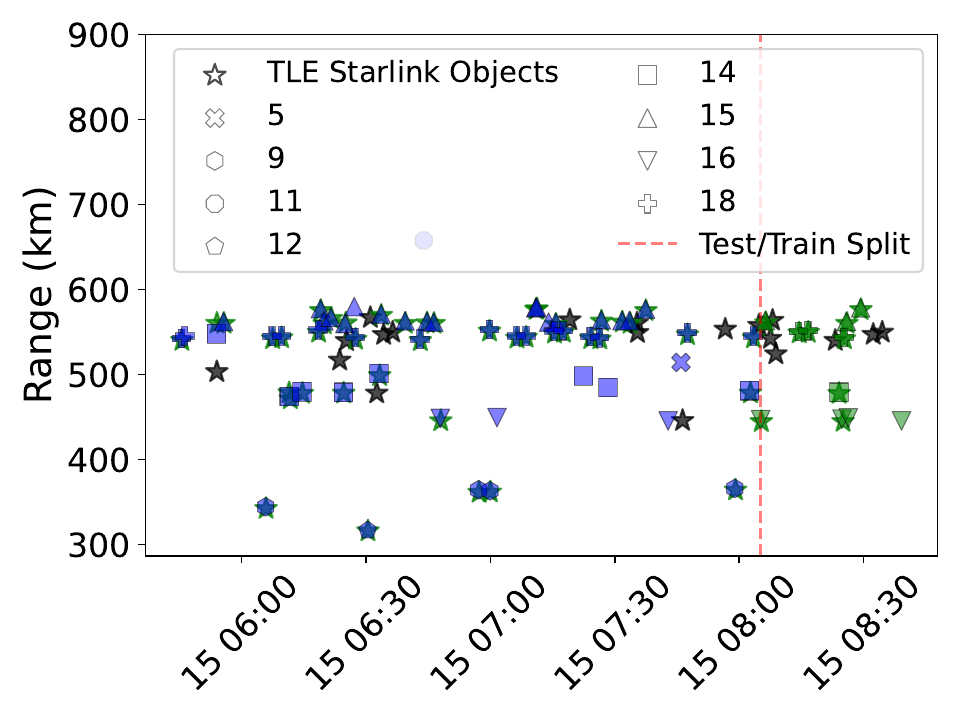}
    \caption{NMFk and XGBoost classifier detections compared with TLE-derived Starlink objects. Star markers denote known Starlink objects from TLE data across the full observation window. The dashed vertical line separates the first 80\% of the data, used for NMFk clustering and SME interpretation, from the final 20\%, used as the classifier test window. On both sides of the split, green stars indicate TLE Starlink objects that were detected by the corresponding Starlink-labeled clusters or classifier predictions. In the test window, green cluster markers indicate classifier predictions that agree with the NMFk cluster labels. Cluster marker shapes correspond to the NMFk cluster identifiers shown in the legend; see Figure~\ref{cluster_spectra} for the cluster depiction.}
    \label{fig:tle_comparison}
\end{figure}

The results in this section address two related questions. First, how well do the NMFk clusters, after SME interpretation, align with known tracked Starlink objects? Second, how well does the XGBoost classifier reproduce the NMFk-based Starlink labels in the held-out test window, and how many known Starlink satellites does it recover?

Figure~\ref{fig:tle_comparison} summarizes both evaluations. The full observation window is overlaid with TLE-derived Starlink objects, shown as star markers. The vertical dashed line separates the first 80\% of the data, used for NMFk clustering and SME interpretation, from the final 20\%, used to evaluate the classifier. In the training window, clusters that were interpreted by the SME as ``Starlink'' are compared against the known TLE Starlink objects. A green star indicates that a known Starlink object was detected by one of these SME-labeled NMFk clusters. Using this criterion, the NMFk+SME workflow detected 44 of the 57 known Starlink satellites in the training portion of the data.

The right side of Figure~\ref{fig:tle_comparison} evaluates the classifier on the held-out 20\% test window. Here, green cluster markers indicate cases where the classifier prediction agrees with the corresponding NMFk cluster label. The classifier showed perfect agreement with the SME-labeled NMFk Starlink clusters in the test data, indicating that it successfully learned to reproduce the NMFk-based labeling behavior. We then compared the classifier-predicted Starlink clusters against the known TLE Starlink objects in the same test window. Green stars indicate TLE Starlink objects recovered by the classifier predictions. Under this evaluation, the classifier detected 9 of the 15 expected Starlink satellites in the test window.

All spectra assigned to the SME-interpreted ``Starlink'' clusters were reviewed and found to contain clear satellite signatures. In several cases, detections did not coincide with a known Starlink object in the TLE overlay. For these cases, we identified other non-Starlink satellites with similar range and Doppler-frequency behavior in the same time window, although these additional objects are not shown in Figure~\ref{fig:tle_comparison}. The missed Starlink detections and apparent non-Starlink detections suggest that future work should focus on improving the clustering resolution and label interpretation process. We also plan to benchmark this workflow against a conventional SNR-threshold detector to distinguish the contribution of latent factorization from that of basic energy-based detection.



\section{Conclusion}
\label{sec:conclusion}
Detecting, observing, and characterizing space objects in large datasets requires prior knowledge and extensive data labeling, both of which are often scarce. To develop a meaningful detection method for sparsely observed satellites, we proposed a gradient-boosted classifier trained on NMFk clustering results derived from unlabeled RF spectral data. In this paper, we introduced a semi-supervised detection pipeline that includes raw RF data processing, NMFk cluster generation, SME labeling, XGBoost classification, and TLE-derived ground-truth validation. This workflow is capable of detecting and classifying spectral signatures of space objects with a high degree of accuracy.

The latent feature detection used in NMFk clustering provides an unsupervised mechanism for representing meaningful underlying physical phenomena in RF spectral data. These unsupervised NMFk-generated patterns can be leveraged by subject matter experts to identify physical events in unlabeled RF data that would otherwise require extensive expert-level review of every sample. XGBoost classifiers trained on these semi-supervised, SME-selected clusters can then be efficiently applied to future RF spectral data to detect the same signatures. Meaning can be assigned to future data by linking each predicted cluster back to the SME-defined physical interpretation of that cluster, thereby automating aspects of detection, observation, and characterization.

Our proof-of-concept application of NMFk to unlabeled RF spectral data demonstrates a promising methodology for classifying and detecting space objects, while also leaving room for improved accuracy. Ongoing work is addressing the potential for false-negative detections buried in clusters labeled as not containing objects. To address this issue, we are actively developing and testing a hierarchical approach that separates first-round detections from background spectra and then re-investigates those clusters using a second round of NMFk. Early results show the potential to detect previously missed objects and increase the likelihood of detection. Additionally, a multi-channel approach will leverage ultra-high-frequency (UHF) and very-high-frequency (VHF) channels within the ALTAIR data, using both the principal-polarization and orthogonal-polarization detection modalities. This multi-channel approach aims to use different frequencies and polarization states to exploit geometric and other physical properties of space objects, thereby improving classification accuracy and precision. Ultimately, we aim to expand this classification and validation framework to characterize a wider range of space objects using the same methodology.

More broadly, our results suggest that semi-supervised classification workflows are a practical solution for RF event classification when labeled data are scarce, observational conditions are evolving, and interpretability is crucial.

\section*{Acknowledgment}
This manuscript has been approved for unlimited release and has been assigned LA-UR-26-24144. This research is based upon work supported in part by the Office of the Director of National Intelligence (ODNI), Intelligence Advanced Research Projects Activity (IARPA). The views and conclusions contained herein are those of the authors and should not be interpreted as necessarily representing the official policies, either expressed or implied, of ODNI, IARPA, or the U.S. Government. The U.S. Government is authorized to reproduce and distribute reprints for governmental purposes notwithstanding any copyright annotation therein. 
LANL is operated by Triad National Security, LLC, for the National Nuclear Security Administration of the U.S. Department of Energy (Contract No. 89233218CNA000001). This research used resources provided by the Los Alamos National Laboratory Institutional Computing Program, which is supported by the U.S. Department of Energy National Nuclear Security Administration under Contract No. 89233218CNA000001.

\bibliographystyle{IEEEtran}
\bibliography{References}

@article{bennett2025orbital,
  title={Orbital debris requires prevention and mitigation across the satellite life cycle},
  author={Bennett, Mia M},
  journal={Communications Engineering},
  volume={4},
  number={1},
  pages={95},
  year={2025},
  publisher={Nature Publishing Group UK London}
}

@misc{TELF,
  title        = {{Tensor Extraction of Latent Features (T-ELF)}},
  author       = {Eren, Maksim and Solovyev, Nick and Barron, Ryan and Bhattarai, Manish and Truong, Duc and Boureima, Ismael and Skau, Erik and Rasmussen, Kim \O. and Alexandrov, Boian},
  year         = 2023,
  month        = oct,
  doi          = {10.5281/zenodo.10257897},
  url          = {https://github.com/lanl/T-ELF}
}

@article{Redmon2015YouOL,
  title={You Only Look Once: Unified, Real-Time Object Detection},
  author={Joseph Redmon and Santosh Kumar Divvala and Ross B. Girshick and Ali Farhadi},
  journal={2016 IEEE Conference on Computer Vision and Pattern Recognition (CVPR)},
  year={2015},
  pages={779-788},
  url={https://api.semanticscholar.org/CorpusID:206594738}
}

@inproceedings{10.1145/2939672.2939785,
author = {Chen, Tianqi and Guestrin, Carlos},
title = {XGBoost: A Scalable Tree Boosting System},
year = {2016},
isbn = {9781450342322},
publisher = {Association for Computing Machinery},
address = {New York, NY, USA},
url = {https://doi.org/10.1145/2939672.2939785},
doi = {10.1145/2939672.2939785},
booktitle = {Proceedings of the 22nd ACM SIGKDD International Conference on Knowledge Discovery and Data Mining},
pages = {785–794},
numpages = {10},
location = {San Francisco, California, USA},
series = {KDD '16}
}

@article{zhu2025cross,
  title={Cross-Domain Object Detection with Hierarchical Multi-Scale Domain Adaptive YOLO},
  author={Zhu, Sihan and Zhu, Peipei and Wu, Yuan and Qiao, Wensheng},
  journal={Sensors},
  volume={25},
  number={17},
  pages={5363},
  year={2025},
  publisher={MDPI}
}

@book{national2011limiting,
  title={Limiting future collision risk to spacecraft: an assessment of NASA's meteoroid and orbital debris programs},
  author={National Research Council and Division on Engineering and Physical Sciences and Aeronautics and Space Engineering Board and Committee for the Assessment of NASA's Orbital Debris Programs},
  year={2011},
  publisher={National Academies Press}
}

@article{schildknecht2007optical,
  title={Optical surveys for space debris},
  author={Schildknecht, Thomas},
  journal={The Astronomy and Astrophysics Review},
  volume={14},
  number={1},
  pages={41--111},
  year={2007},
  publisher={Springer}
}

@article{hickson2018fast,
  title={A fast algorithm for the detection of faint orbital debris tracks in optical images},
  author={Hickson, P},
  journal={Advances in space research},
  volume={62},
  number={11},
  pages={3078--3085},
  year={2018},
  publisher={Elsevier}
}

@article{foster2005detection,
  title={Detection of small radar cross-section orbital debris with the Haystack radar},
  author={Foster, JL and Benbrook, JR and Stansbery, EG},
  journal={Advances in Space Research},
  volume={35},
  number={7},
  pages={1210--1213},
  year={2005},
  publisher={Elsevier}
}

@article{mehrholz2002detecting,
  title={Detecting, tracking and imaging space debris},
  author={Mehrholz, Dieter and Leushacke, Ludger and Flury, W and Jehn, Rudiger and Klinkrad, H and Landgraf, M},
  journal={ESA Bulletin(0376-4265)},
  number={109},
  pages={128--134},
  year={2002}
}

@article{prabu2020development,
  title={The development of non-coherent passive radar techniques for space situational awareness with the Murchison Widefield Array},
  author={Prabu, Steve and Hancock, Paul J and Zhang, Xiang and Tingay, Steven J},
  journal={Publications of the Astronomical Society of Australia},
  volume={37},
  pages={e010},
  year={2020},
  publisher={Cambridge University Press}
}

@article{massimi2024deep,
  title={Deep learning-based space debris detection for space situational awareness: A feasibility study applied to the radar processing},
  author={Massimi, Federica and Ferrara, Pasquale and Petrucci, Roberto and Benedetto, Francesco},
  journal={IET Radar, Sonar \& Navigation},
  volume={18},
  number={4},
  pages={635--648},
  year={2024},
  publisher={Wiley Online Library}
}

@article{zhang2025semi,
  title={A semi-supervised object detection method for close range detection of spacecraft and space debris},
  author={Zhang, Huan and Zhang, Yang and Feng, Qingjuan and Zhang, Kebei},
  journal={International Journal of Aeronautical and Space Sciences},
  volume={26},
  number={2},
  pages={773--784},
  year={2025},
  publisher={Springer}
}

@article{eren2023semi,
  title={Semi-supervised classification of malware families under extreme class imbalance via hierarchical non-negative matrix factorization with automatic model selection},
  author={Eren, Maksim E and Bhattarai, Manish and Joyce, Robert J and Raff, Edward and Nicholas, Charles and Alexandrov, Boian S},
  journal={ACM Transactions on Privacy and Security},
  volume={26},
  number={4},
  pages={1--27},
  year={2023},
  publisher={ACM New York, NY}
}

@book{chapelle2006semi,
  editor    = {Chapelle, Olivier and Sch{\"o}lkopf, Bernhard and Zien, Alexander},
  title     = {Semi-Supervised Learning},
  publisher = {MIT Press},
  year      = {2006},
  doi       = {10.7551/mitpress/9780262033589.001.0001}
}

@article{lee1999learning,
  title={Learning the parts of objects by non-negative matrix factorization},
  author={Lee, Daniel D and Seung, H Sebastian},
  journal={Nature},
  volume={401},
  number={6755},
  pages={788--791},
  year={1999}
}

@article{bishop1999bayesian,
  title={Bayesian pca},
  author={Bishop, Christopher M},
  journal={Advances in neural information processing systems},
  pages={382--388},
  year={1999},
  publisher={MIT; 1998}
}

@article{mackay1994bayesian,
  title={Bayesian nonlinear modeling for the prediction competition},
  author={MacKay, David JC},
  journal={ASHRAE transactions},
  volume={100},
  number={2},
  pages={1053--1062},
  year={1994},
  publisher={Citeseer}
}

@inproceedings{fevotte2009nonnegative,
  title={Nonnegative matrix factorizations as probabilistic inference in composite models},
  author={F{\'e}votte, C{\'e}dric and Cemgil, A Taylan},
  booktitle={17th European Signal Processing Conference},
  pages={1913--1917},
  year={2009}
}

@inproceedings{morup2009tuning,
  title={Tuning pruning in sparse non-negative matrix factorization},
  author={M{\o}rup, Morten and Hansen, Lars Kai},
  booktitle={2009 17th European Signal Processing Conference},
  pages={1923--1927},
  year={2009},
  organization={IEEE}
}

@article{tan2012automatic,
  title={Automatic relevance determination in nonnegative matrix factorization with the/spl beta/-divergence},
  author={Tan, Vincent YF and F{\'e}votte, C{\'e}dric},
  journal={IEEE Transactions on Pattern Analysis and Machine Intelligence},
  volume={35},
  number={7},
  pages={1592--1605},
  year={2012},
  publisher={IEEE}
}

@article{brunet2004metagenes,
  title={Metagenes and molecular pattern discovery using matrix factorization},
  author={Brunet, Jean-Philippe and Tamayo, Pablo and Golub, Todd R and Mesirov, Jill P},
  journal={Proceedings of the national academy of sciences},
  volume={101},
  number={12},
  pages={4164--4169},
  year={2004},
  publisher={National Acad Sciences}
}

@misc{alexandrov2020patent,
  title={Source identification by non-negative matrix factorization combined with semi-supervised clustering},
  author={Alexandrov, Boian S and Alexandrov, Ludmil B and Iliev, Filip L and Stanev, Valentin G and Vesselinov, Velimir V},
  year={2020},
  journal={US Patent S10,776,718},
  publisher={United States Patent and Trademark Office (USPTO)},
  note={US Patent 10,776,718}
}

@article{alexandrov2013signatures,
author = {{L. B. Alexandrov et al.}},
title={Signatures of mutational processes in human cancer},
journal={Nature},
year={2013},
month={Aug},
day={01},
volume={500},
number={7463},
pages={415-421},
issn={1476-4687},
doi={10.1038/nature12477},
url={https://doi.org/10.1038/nature12477}
}

@article{alexandrov2020repertoire,
author = {{L. B. Alexandrov et al.}},
title={The repertoire of mutational signatures in human cancer},
journal={Nature},
year={2020},
month={Feb},
day={01},
volume={578},
number={7793},
pages={94-101},
issn={1476-4687},
doi={10.1038/s41586-020-1943-3},
url={https://doi.org/10.1038/s41586-020-1943-3}
}

@article{islam2022uncovering,
author = {{S. M. Ashiqul Islam et. al.}},
title = {Uncovering novel mutational signatures by de novo extraction with SigProfilerExtractor},
journal = {Cell Genomics},
volume = {2},
number = {11},
pages = {100179},
year = {2022},
issn = {2666-979X},
doi = {https://doi.org/10.1016/j.xgen.2022.100179},
url = {https://www.sciencedirect.com/science/article/pii/S2666979X22001240}
}

@article{nebgen2021neural,
  title={A neural network for determination of latent dimensionality in non-negative matrix factorization},
  author={Nebgen, Benjamin T and Vangara, Raviteja and Hombrados-Herrera, Miguel A and Kuksova, Svetlana and Alexandrov, Boian S},
  journal={Machine Learning: Science and Technology},
  volume={2},
  number={2},
  pages={025012},
  year={2021},
  publisher={IOP Publishing}
}

@inproceedings{vangara2020semantic,
  title={Semantic nonnegative matrix factorization with automatic model determination for topic modeling},
  author={Vangara, Raviteja and Skau, Erik and Chennupati, Gopinath and Djidjev, Hristo and Tierney, Thomas and Smith, James P and Bhattarai, Manish and Stanev, Valentin G and Alexandrov, Boian S},
  booktitle={2020 19th IEEE International Conference on Machine Learning and Applications (ICMLA)},
  pages={328--335},
  year={2020},
  organization={IEEE}
}

@article{ROUSSEEUW198753,
title = {Silhouettes: A graphical aid to the interpretation and validation of cluster analysis},
journal = {Journal of Computational and Applied Mathematics},
volume = {20},
pages = {53-65},
year = {1987},
issn = {0377-0427},
doi = {https://doi.org/10.1016/0377-0427(87)90125-7},
url = {https://www.sciencedirect.com/science/article/pii/0377042787901257},
author = {Peter J. Rousseeuw}
}

@article{Dougherty1961,
    author = {Dougherty, J. P. and Farley, D. T.},
    title = {A theory of incoherent scattering of radio waves by a plasma},
    journal = {Proceedings of the Royal Society of London. A. Mathematical and Physical Sciences},
    volume = {259},
    number = {1296},
    pages = {79-99},
    year = {1961},
    month = {01},
    issn = {0080-4630},
    doi = {10.1098/rspa.1960.0212},
    url = {https://doi.org/10.1098/rspa.1960.0212},
    eprint = {https://royalsocietypublishing.org/rspa/article-pdf/259/1296/79/52523/rspa.1960.0212.pdf},
}

@article{vangara2021finding,
  title={Finding the Number of Latent Topics with Semantic Non-negative Matrix Factorization},
  author={Vangara, Raviteja and Bhattarai, Manish and Skau, Erik and Chennupati, Gopinath and Djidjev, Hristo and Tierney, Thomas and Smith, James P and Stanev, Valentin G and Alexandrov, Boian S},
  journal={IEEE Access},
  year={2021},
  publisher={IEEE}
}

@Inbook{Haynes2013,
author="Haynes, Winston",
editor="Dubitzky, Werner
and Wolkenhauer, Olaf
and Cho, Kwang-Hyun
and Yokota, Hiroki",
title="Wilcoxon Rank Sum Test",
bookTitle="Encyclopedia of Systems Biology",
year="2013",
publisher="Springer New York",
address="New York, NY",
pages="2354--2355",
isbn="978-1-4419-9863-7",
doi="10.1007/978-1-4419-9863-7_1185"
}

@article{Akbari2017,
  title={Incoherent Scatter Plasma Lines: Observations and Applications},
  author={Hassanali Akbari and Asti Bhatt and Cesar La Hoz and Joshua L. Semeter},
  journal={Space Science Reviews},
  year={2017},
  volume={212},
  pages={249-294},
  url={https://api.semanticscholar.org/CorpusID:126219370}
}

@misc{sintra,
  author= {Truitt, Alexis},
  title = { SINTRA Space Debris Identification and Tracking},
  howpublished = {\url{https://www.iarpa.gov/research-programs/sintra}},
  note = {Accessed: 2026-04-30}
}

@article{Hysell2020,
author = {Hysell, D. L. and Rao, S. and Groves, K. M. and Larsen, M. F.},
title = {Radar Investigation of Postsunset Equatorial Ionospheric Instability Over Kwajalein During Project WINDY},
journal = {Journal of Geophysical Research: Space Physics},
volume = {125},
number = {6},
pages = {e2020JA027997},
doi = {https://doi.org/10.1029/2020JA027997},
url = {https://agupubs.onlinelibrary.wiley.com/doi/abs/10.1029/2020JA027997},
eprint = {https://agupubs.onlinelibrary.wiley.com/doi/pdf/10.1029/2020JA027997},
note = {e2020JA027997 10.1029/2020JA027997},
year = {2020}
}

@article{Stone2000,
author ={Melvin L. Stone and Gerald P. Banner},
title = {Radars for the detection and tracking of ballistic missiles, satellites, and planets.},
journal={Lincoln Laboratory Journal}, volume={12(2)},
pages={217–244},
year={2000},
url={https://archive.ll.mit.edu/publications/journal/pdf/vol12_no2/12_2detectsatellitiesplanets.pdf}
}

@misc{spacetrack,
author=USSPACECOM,
title={space-track.org},
url={https://www.space-track.org},
accessed={2026-04-13}
}

@misc{celestrak,
author={Kelso, T.S.},
title={CelesTrak},
url={https://celestrak.org/},
accessed={2026-04-13}
}

\vspace{12pt}

\end{document}